\documentclass[conference]{IEEEtran}
\IEEEoverridecommandlockouts
\usepackage{amsmath,amsfonts} 
\usepackage{amssymb, amsthm}
\usepackage{graphicx}
\usepackage{algorithm}
\usepackage{algpseudocode}
\usepackage{verbatim}
\usepackage{comment}
\usepackage{geometry}
\usepackage{afterpage}

\makeatletter
\let\MYcaption\@makecaption
\makeatother
\usepackage[font=footnotesize]{subcaption}
\makeatletter
\let\@makecaption\MYcaption
\makeatother

\usepackage{booktabs}
\usepackage{color}
\usepackage{microtype}
\usepackage{balance}
\usepackage{cite}

\usepackage{amssymb}
\usepackage{amsfonts}
\usepackage{mathrsfs}
\usepackage{xspace}
\usepackage{bm}
\usepackage{upgreek}

\newcommand{\safemath}[2]{\newcommand{#1}{\ensuremath{#2}\xspace}}

\safemath{\bma}{\mathbf{a}}
\safemath{\bmb}{\mathbf{b}}
\safemath{\bmc}{\mathbf{c}}
\safemath{\bmd}{\mathbf{d}}
\safemath{\bme}{\mathbf{e}}
\safemath{\bmf}{\mathbf{f}}
\safemath{\bmg}{\mathbf{g}}
\safemath{\bmh}{\mathbf{h}}
\safemath{\bmi}{\mathbf{i}}
\safemath{\bmj}{\mathbf{j}}
\safemath{\bmk}{\mathbf{k}}
\safemath{\bml}{\mathbf{l}}
\safemath{\bmm}{\mathbf{m}}
\safemath{\bmn}{\mathbf{n}}
\safemath{\bmo}{\mathbf{o}}
\safemath{\bmp}{\mathbf{p}}
\safemath{\bmq}{\mathbf{q}}
\safemath{\bmr}{\mathbf{r}}
\safemath{\bms}{\mathbf{s}}
\safemath{\bmt}{\mathbf{t}}
\safemath{\bmu}{\mathbf{u}}
\safemath{\bmv}{\mathbf{v}}
\safemath{\bmw}{\mathbf{w}}
\safemath{\bmx}{\mathbf{x}}
\safemath{\bmy}{\mathbf{y}}
\safemath{\bmz}{\mathbf{z}}
\safemath{\bmzero}{\mathbf{0}}
\safemath{\bmone}{\mathbf{1}}

\bmdefine{\biad}{a}
\bmdefine{\bibd}{b}
\bmdefine{\bicd}{c}
\bmdefine{\bidd}{d}
\bmdefine{\bied}{e}
\bmdefine{\bifd}{f}
\bmdefine{\bigd}{g}
\bmdefine{\bihd}{h}
\bmdefine{\biid}{i}
\bmdefine{\bijd}{j}
\bmdefine{\bikd}{k}
\bmdefine{\bild}{l}
\bmdefine{\bimd}{m}
\bmdefine{\bind}{n}
\bmdefine{\biod}{o}
\bmdefine{\bipd}{p}
\bmdefine{\biqd}{q}
\bmdefine{\bird}{r}
\bmdefine{\bisd}{s}
\bmdefine{\bitd}{t}
\bmdefine{\biud}{u}
\bmdefine{\bivd}{v}
\bmdefine{\biwd}{w}
\bmdefine{\bixd}{x}
\bmdefine{\biyd}{y}
\bmdefine{\bizd}{z}

\bmdefine{\bixid}{\xi}
\bmdefine{\bilambdad}{\lambda}
\bmdefine{\bimud}{\mu}
\bmdefine{\bithetad}{\theta}
\bmdefine{\biphid}{\phi}
\bmdefine{\bideltad}{\delta}

\safemath{\bmia}{\biad}
\safemath{\bmib}{\bibd}
\safemath{\bmic}{\bicd}
\safemath{\bmid}{\bidd}
\safemath{\bmie}{\bied}
\safemath{\bmif}{\bifd}
\safemath{\bmig}{\bigd}
\safemath{\bmih}{\bihd}
\safemath{\bmii}{\biid}
\safemath{\bmij}{\bijd}
\safemath{\bmik}{\bikd}
\safemath{\bmil}{\bild}
\safemath{\bmim}{\bimd}
\safemath{\bmin}{\bind}
\safemath{\bmio}{\biod}
\safemath{\bmip}{\bipd}
\safemath{\bmiq}{\biqd}
\safemath{\bmir}{\bird}
\safemath{\bmis}{\bisd}
\safemath{\bmit}{\bitd}
\safemath{\bmiu}{\biud}
\safemath{\bmiv}{\bivd}
\safemath{\bmiw}{\biwd}
\safemath{\bmix}{\bixd}
\safemath{\bmiy}{\biyd}
\safemath{\bmiz}{\bizd}

\safemath{\bmxi}{\bixid}
\safemath{\bmlambda}{\bilambdad}
\safemath{\bmmu}{\bimud}
\safemath{\bmtheta}{\bithetad}
\safemath{\bmphi}{\biphid}
\safemath{\bmdelta}{\bideltad}

\safemath{\bA}{\mathbf{A}}
\safemath{\bB}{\mathbf{B}}
\safemath{\bC}{\mathbf{C}}
\safemath{\bD}{\mathbf{D}}
\safemath{\bE}{\mathbf{E}}
\safemath{\bF}{\mathbf{F}}
\safemath{\bG}{\mathbf{G}}
\safemath{\bH}{\mathbf{H}}
\safemath{\bI}{\mathbf{I}}
\safemath{\bJ}{\mathbf{J}}
\safemath{\bK}{\mathbf{K}}
\safemath{\bL}{\mathbf{L}}
\safemath{\bM}{\mathbf{M}}
\safemath{\bN}{\mathbf{N}}
\safemath{\bO}{\mathbf{O}}
\safemath{\bP}{\mathbf{P}}
\safemath{\bQ}{\mathbf{Q}}
\safemath{\bR}{\mathbf{R}}
\safemath{\bS}{\mathbf{S}}
\safemath{\bT}{\mathbf{T}}
\safemath{\bU}{\mathbf{U}}
\safemath{\bV}{\mathbf{V}}
\safemath{\bW}{\mathbf{W}}
\safemath{\bX}{\mathbf{X}}
\safemath{\bY}{\mathbf{Y}}
\safemath{\bZ}{\mathbf{Z}}

\safemath{\bZero}{\mathbf{0}}
\safemath{\bOne}{\mathbf{1}}
\safemath{\bDelta}{\mathbf{\Delta}}
\safemath{\bLambda}{\mathbf{\UpLambda}}
\safemath{\bPhi}{\mathbf{\Upphi}}
\safemath{\bSigma}{\mathbf{\Upsigma}}
\safemath{\bOmega}{\mathbf{\Upomega}}
\safemath{\bTheta}{\mathbf{\Uptheta}}

\bmdefine{\biAd}{A}
\bmdefine{\biBd}{B}
\bmdefine{\biCd}{C}
\bmdefine{\biDd}{D}
\bmdefine{\biEd}{E}
\bmdefine{\biFd}{F}
\bmdefine{\biGd}{G}
\bmdefine{\biHd}{H}
\bmdefine{\biId}{I}
\bmdefine{\biJd}{J}
\bmdefine{\biKd}{K}
\bmdefine{\biLd}{L}
\bmdefine{\biMd}{M}
\bmdefine{\biOd}{N}
\bmdefine{\biPd}{O}
\bmdefine{\biQd}{P}
\bmdefine{\biRd}{R}
\bmdefine{\biSd}{S}
\bmdefine{\biTd}{T}
\bmdefine{\biUd}{U}
\bmdefine{\biVd}{V}
\bmdefine{\biWd}{W}
\bmdefine{\biXd}{X}
\bmdefine{\biYd}{Y}
\bmdefine{\biZd}{Z}

\bmdefine{\biDelta}{\Delta}
\bmdefine{\biLambda}{\Lambda}
\bmdefine{\biPhi}{\Phi}
\bmdefine{\biSigma}{\Sigma}
\bmdefine{\biOmega}{\Omega}
\bmdefine{\biTheta}{\Theta}

\safemath{\bimA}{\biAd}
\safemath{\bimB}{\biBd}
\safemath{\bimC}{\biCd}
\safemath{\bimD}{\biDd}
\safemath{\bimE}{\biEd}
\safemath{\bimF}{\biFd}
\safemath{\bimG}{\biGd}
\safemath{\bimH}{\biHd}
\safemath{\bimI}{\biId}
\safemath{\bimJ}{\biJd}
\safemath{\bimK}{\biKd}
\safemath{\bimL}{\biLd}
\safemath{\bimM}{\biMd}
\safemath{\bimN}{\biNd}
\safemath{\bimO}{\biOd}
\safemath{\bimP}{\biPd}
\safemath{\bimQ}{\biQd}
\safemath{\bimR}{\biRd}
\safemath{\bimS}{\biSd}
\safemath{\bimT}{\biTd}
\safemath{\bimU}{\biUd}
\safemath{\bimV}{\biVd}
\safemath{\bimW}{\biWd}
\safemath{\bimX}{\biXd}
\safemath{\bimY}{\biYd}
\safemath{\bimZ}{\biZd}

\safemath{\bimDelta}{\biDelta}
\safemath{\bimLambda}{\biLambda}
\safemath{\bimPhi}{\biPhi}
\safemath{\bimSigma}{\biSigma}
\safemath{\bimOmega}{\biOmega}
\safemath{\bimTheta}{\biTheta}

\safemath{\setA}{\mathcal{A}}
\safemath{\setB}{\mathcal{B}}
\safemath{\setC}{\mathcal{C}}
\safemath{\setD}{\mathcal{D}}
\safemath{\setE}{\mathcal{E}}
\safemath{\setF}{\mathcal{F}}
\safemath{\setG}{\mathcal{G}}
\safemath{\setH}{\mathcal{H}}
\safemath{\setI}{\mathcal{I}}
\safemath{\setJ}{\mathcal{J}}
\safemath{\setK}{\mathcal{K}}
\safemath{\setL}{\mathcal{L}}
\safemath{\setM}{\mathcal{M}}
\safemath{\setN}{\mathcal{N}}
\safemath{\setO}{\mathcal{O}}
\safemath{\setP}{\mathcal{P}}
\safemath{\setQ}{\mathcal{Q}}
\safemath{\setR}{\mathcal{R}}
\safemath{\setS}{\mathcal{S}}
\safemath{\setT}{\mathcal{T}}
\safemath{\setU}{\mathcal{U}}
\safemath{\setV}{\mathcal{V}}
\safemath{\setW}{\mathcal{W}}
\safemath{\setX}{\mathcal{X}}
\safemath{\setY}{\mathcal{Y}}
\safemath{\setZ}{\mathcal{Z}}
\safemath{\emptySet}{\varnothing}

\safemath{\colA}{\mathscr{A}}
\safemath{\colB}{\mathscr{B}}
\safemath{\colC}{\mathscr{C}}
\safemath{\colD}{\mathscr{D}}
\safemath{\colE}{\mathscr{E}}
\safemath{\colF}{\mathscr{F}}
\safemath{\colG}{\mathscr{G}}
\safemath{\colH}{\mathscr{H}}
\safemath{\colI}{\mathscr{I}}
\safemath{\colJ}{\mathscr{J}}
\safemath{\colK}{\mathscr{K}}
\safemath{\colL}{\mathscr{L}}
\safemath{\colM}{\mathscr{M}}
\safemath{\colN}{\mathscr{N}}
\safemath{\colO}{\mathscr{O}}
\safemath{\colP}{\mathscr{P}}
\safemath{\colQ}{\mathscr{Q}}
\safemath{\colR}{\mathscr{R}}
\safemath{\colS}{\mathscr{S}}
\safemath{\colT}{\mathscr{T}}
\safemath{\colU}{\mathscr{U}}
\safemath{\colV}{\mathscr{V}}
\safemath{\colW}{\mathscr{W}}
\safemath{\colX}{\mathscr{X}}
\safemath{\colY}{\mathscr{Y}}
\safemath{\colZ}{\mathscr{Z}}

\safemath{\opA}{\mathbb{A}}
\safemath{\opB}{\mathbb{B}}
\safemath{\opC}{\mathbb{C}}
\safemath{\opD}{\mathbb{D}}
\safemath{\opE}{\mathbb{E}}
\safemath{\opF}{\mathbb{F}}
\safemath{\opG}{\mathbb{G}}
\safemath{\opH}{\mathbb{H}}
\safemath{\opI}{\mathbb{I}}
\safemath{\opJ}{\mathbb{J}}
\safemath{\opK}{\mathbb{K}}
\safemath{\opL}{\mathbb{L}}
\safemath{\opM}{\mathbb{M}}
\safemath{\opN}{\mathbb{N}}
\safemath{\opO}{\mathbb{O}}
\safemath{\opP}{\mathbb{P}}
\safemath{\opQ}{\mathbb{Q}}
\safemath{\opR}{\mathbb{R}}
\safemath{\opS}{\mathbb{S}}
\safemath{\opT}{\mathbb{T}}
\safemath{\opU}{\mathbb{U}}
\safemath{\opV}{\mathbb{V}}
\safemath{\opW}{\mathbb{W}}
\safemath{\opX}{\mathbb{X}}
\safemath{\opY}{\mathbb{Y}}
\safemath{\opZ}{\mathbb{Z}}
\safemath{\opZero}{\mathbb{O}}
\safemath{\identityop}{\opI}

\safemath{\veca}{\bma}
\safemath{\vecb}{\bmb}
\safemath{\vecc}{\bmc}
\safemath{\vecd}{\bmd}
\safemath{\vece}{\bme}
\safemath{\vecf}{\bmf}
\safemath{\vecg}{\bmg}
\safemath{\vech}{\bmh}
\safemath{\veci}{\bmi}
\safemath{\vecj}{\bmj}
\safemath{\veck}{\bmk}
\safemath{\vecl}{\bml}
\safemath{\vecm}{\bmm}
\safemath{\vecn}{\bmn}
\safemath{\veco}{\bmo}
\safemath{\vecp}{\bmp}
\safemath{\vecq}{\bmq}
\safemath{\vecr}{\bmr}
\safemath{\vecs}{\bms}
\safemath{\vect}{\bmt}
\safemath{\vecu}{\bmu}
\safemath{\vecv}{\bmv}
\safemath{\vecw}{\bmw}
\safemath{\vecx}{\bmx}
\safemath{\vecy}{\bmy}
\safemath{\vecz}{\bmz}

\safemath{\veczero}{\bmzero}
\safemath{\vecone}{\bmone}
\safemath{\vecxi}{\bmxi}
\safemath{\veclambda}{\bmlambda}
\safemath{\vecmu}{\bmmu}
\safemath{\vectheta}{\bmtheta}
\safemath{\vecphi}{\bmphi}
\safemath{\vecdelta}{\bmdelta}

\safemath{\matA}{\bA}
\safemath{\matB}{\bB}
\safemath{\matC}{\bC}
\safemath{\matD}{\bD}
\safemath{\matE}{\bE}
\safemath{\matF}{\bF}
\safemath{\matG}{\bG}
\safemath{\matH}{\bH}
\safemath{\matI}{\bI}
\safemath{\matJ}{\bJ}
\safemath{\matK}{\bK}
\safemath{\matL}{\bL}
\safemath{\matM}{\bM}
\safemath{\matN}{\bN}
\safemath{\matO}{\bO}
\safemath{\matP}{\bP}
\safemath{\matQ}{\bQ}
\safemath{\matR}{\bR}
\safemath{\matS}{\bS}
\safemath{\matT}{\bT}
\safemath{\matU}{\bU}
\safemath{\matV}{\bV}
\safemath{\matW}{\bW}
\safemath{\matX}{\bX}
\safemath{\matY}{\bY}
\safemath{\matZ}{\bZ}
\safemath{\matzero}{\bmzero}

\safemath{\matDelta}{\bDelta}
\safemath{\matLambda}{\bLambda}
\safemath{\matPhi}{\bPhi}
\safemath{\matSigma}{\bSigma}
\safemath{\matOmega}{\bOmega}
\safemath{\matTheta}{\bTheta}

\safemath{\matidentity}{\matI}
\safemath{\matone}{\matO}

\safemath{\rnda}{A}
\safemath{\rndb}{B}
\safemath{\rndc}{C}
\safemath{\rndd}{D}
\safemath{\rnde}{E}
\safemath{\rndf}{F}
\safemath{\rndg}{G}
\safemath{\rndh}{H}
\safemath{\rndi}{I}
\safemath{\rndj}{J}
\safemath{\rndk}{K}
\safemath{\rndl}{L}
\safemath{\rndm}{M}
\safemath{\rndn}{N}
\safemath{\rndo}{O}
\safemath{\rndp}{P}
\safemath{\rndq}{Q}
\safemath{\rndr}{R}
\safemath{\rnds}{S}
\safemath{\rndt}{T}
\safemath{\rndu}{U}
\safemath{\rndv}{V}
\safemath{\rndw}{W}
\safemath{\rndx}{X}
\safemath{\rndy}{Y}
\safemath{\rndz}{Z}

\safemath{\rveca}{\bimA}
\safemath{\rvecb}{\bimB}
\safemath{\rvecc}{\bimC}
\safemath{\rvecd}{\bimD}
\safemath{\rvece}{\bimE}
\safemath{\rvecf}{\bimF}
\safemath{\rvecg}{\bimG}
\safemath{\rvech}{\bimH}
\safemath{\rveci}{\bimI}
\safemath{\rvecj}{\bimJ}
\safemath{\rveck}{\bimK}
\safemath{\rvecl}{\bimL}
\safemath{\rvecm}{\bimM}
\safemath{\rvecn}{\bimN}
\safemath{\rveco}{\bomO}
\safemath{\rvecp}{\bimP}
\safemath{\rvecq}{\bimQ}
\safemath{\rvecr}{\bimR}
\safemath{\rvecs}{\bimS}
\safemath{\rvect}{\bimT}
\safemath{\rvecu}{\bimU}
\safemath{\rvecv}{\bimV}
\safemath{\rvecw}{\bimW}
\safemath{\rvecx}{\bimX}
\safemath{\rvecy}{\bimY}
\safemath{\rvecz}{\bimZ}

\safemath{\rvecxi}{\bmxi}
\safemath{\rveclambda}{\bmlambda}
\safemath{\rvecmu}{\bmmu}
\safemath{\rvectheta}{\bmtheta}
\safemath{\rvecphi}{\bmphi}

\safemath{\rmatA}{\bimA}
\safemath{\rmatB}{\bimB}
\safemath{\rmatC}{\bimC}
\safemath{\rmatD}{\bimD}
\safemath{\rmatE}{\bimE}
\safemath{\rmatF}{\bimF}
\safemath{\rmatG}{\bimG}
\safemath{\rmatH}{\bimH}
\safemath{\rmatI}{\bimI}
\safemath{\rmatJ}{\bimJ}
\safemath{\rmatK}{\bimK}
\safemath{\rmatL}{\bimL}
\safemath{\rmatM}{\bimM}
\safemath{\rmatN}{\bimN}
\safemath{\rmatO}{\bimO}
\safemath{\rmatP}{\bimP}
\safemath{\rmatQ}{\bimQ}
\safemath{\rmatR}{\bimR}
\safemath{\rmatS}{\bimS}
\safemath{\rmatT}{\bimT}
\safemath{\rmatU}{\bimU}
\safemath{\rmatV}{\bimV}
\safemath{\rmatW}{\bimW}
\safemath{\rmatX}{\bimX}
\safemath{\rmatY}{\bimY}
\safemath{\rmatZ}{\bimZ}

\safemath{\rmatDelta}{\bimDelta}
\safemath{\rmatLambda}{\bimLambda}
\safemath{\rmatPhi}{\bimPhi}
\safemath{\rmatSigma}{\bimSigma}
\safemath{\rmatOmega}{\bimOmega}
\safemath{\rmatTheta}{\bimTheta}

\usepackage{amssymb}
\usepackage{amsfonts}
\usepackage{mathrsfs}
\usepackage{xspace}
\usepackage{bm}
\usepackage{fancyref}
\usepackage{textcomp}

\usepackage{multirow}
\usepackage{stmaryrd}

\newenvironment{textbmatrix}{	\setlength{\arraycolsep}{2.5pt}%
								\big[\begin{matrix}}{\end{matrix}\big]%
								\raisebox{0.08ex}{\vphantom{M}}}

\def\be{\begin{equation}}
\def\ee{\end{equation}}
\def\een{\nonumber \end{equation}}
\def\mat{\begin{bmatrix}}
\def\emat{\end{bmatrix}}
\def\btm{\begin{textbmatrix}}
\def\etm{\end{textbmatrix}}

\def\ba#1\ea{\begin{align}#1\end{align}}
\def\bas#1\eas{\begin{align*}#1\end{align*}}
\def\bs#1\es{\begin{split}#1\end{split}} 
\def\bg#1\eg{\begin{gather}#1\end{gather}}
\def\bml#1\eml{\begin{multline}#1\end{multline}}
\def\bi#1\ei{\begin{itemize}#1\end{itemize}}

\safemath{\dirac}{\delta}					
\safemath{\krond}{\dirac}					

\safemath{\upto}{\uparrow}
\safemath{\downto}{\downarrow}
\safemath{\iu}{j}							
\safemath{\ev}{\lambda}						
\safemath{\hilseqspace}{l^{2}}				
\newcommand{\banachfunspace}[1]{\setL^{#1}}	
\safemath{\hilfunspace}{\banachfunspace{2}}	

\safemath{\SNR}{\textsf{SNR}} 				
\safemath{\PAR}{\textsf{PAR}} 				
\safemath{\No}{N_0}							
\safemath{\Es}{E_s}							
\safemath{\Eb}{E_b}							
\safemath{\EbNo}{\frac{\Eb}{\No}}
\safemath{\EsNo}{\frac{\Es}{\No}}

\DeclareMathOperator{\CHop}{\ensuremath{\opH}} 
\safemath{\tvir}{\rndh_{\CHop}}				
\safemath{\tvtf}{\rndl_{\CHop}}				
\safemath{\spf}{\rnds_{\CHop}}				
\safemath{\bff}{H_{\CHop}}					

\safemath{\ircf}{r_{h}}						
\safemath{\tftvcf}{r_{s}}					
\safemath{\tfcf}{r_{l}}						
\safemath{\bfcf}{r_{H}}						

\safemath{\tcorr}{c_h}						
\safemath{\scf}{c_{s}}						
\safemath{\tfcorr}{c_{l}}					
\safemath{\fcorr}{c_{H}}						

\safemath{\mi}{I}							
\safemath{\capacity}{C}						

\safemath{\normal}{\mathcal{N}}			
\safemath{\jpg}{\mathcal{CN}}			
\safemath{\mchain}{\leftrightarrow}		

\safemath{\dB}{\,\mathrm{dB}}
\safemath{\dBm}{\,\mathrm{dBm}}
\safemath{\Hz}{\,\mathrm{Hz}}
\safemath{\kHz}{\,\mathrm{kHz}}
\safemath{\MHz}{\,\mathrm{MHz}}
\safemath{\GHz}{\,\mathrm{GHz}}
\safemath{\s}{\,\mathrm{s}}
\safemath{\ms}{\,\mathrm{ms}}
\safemath{\mus}{\,\mathrm{\text{\textmu}s}}
\safemath{\ns}{\,\mathrm{ns}}
\safemath{\ps}{\,\mathrm{ps}}
\safemath{\meter}{\,\mathrm{m}}
\safemath{\mm}{\,\mathrm{mm}}
\safemath{\cm}{\,\mathrm{cm}}
\safemath{\m}{\,\mathrm{m}}
\safemath{\W}{\,\mathrm{W}}
\safemath{\mW}{\, \mathrm{mW}}
\safemath{\J}{\,\mathrm{J}}
\safemath{\K}{\,\mathrm{K}}
\safemath{\bit}{\,\mathrm{bit}}
\safemath{\nat}{\,\mathrm{nat}}

\safemath{\define}{\triangleq}			

\safemath{\equivalent}{\sim}
\safemath{\distas}{\sim}					
\safemath{\sdiff}{\Delta}				

\safemath{\reals}{\mathbb{R}}
\safemath{\positivereals}{\reals_{+}}
\safemath{\integers}{\mathbb{Z}}
\safemath{\posint}{\integers_{+}}
\safemath{\naturals}{\mathbb{N}}
\safemath{\posnaturals}{\naturals_{+}}
\safemath{\complexset}{\mathbb{C}}
\safemath{\rationals}{\mathbb{Q}}

\newcommand*{\fancyrefapplabelprefix}{app}		
\newcommand*{\fancyrefthmlabelprefix}{thm}		
\newcommand*{\fancyreflemlabelprefix}{lem}		
\newcommand*{\fancyrefcorlabelprefix}{cor}		
\newcommand*{\fancyrefdeflabelprefix}{def}		
\newcommand*{\fancyrefproplabelprefix}{prop}	
\newcommand*{\fancyrefobslabelprefix}{obs}		
\newcommand*{\fancyrefalglabelprefix}{alg}		
\newcommand*{\fancyrefasmlabelprefix}{asm}	    
\newcommand*{\fancyrefasmslabelprefix}{asms}	    
\newcommand*{\fancyreftbllabelprefix}{tbl}	    
\newcommand*{\fancyreftremabelprefix}{rem}	    

\frefformat{vario}{\fancyrefseclabelprefix}{Section~#1}
\frefformat{vario}{\fancyrefthmlabelprefix}{Theorem~#1}
\frefformat{vario}{\fancyreflemlabelprefix}{Lemma~#1}
\frefformat{vario}{\fancyrefcorlabelprefix}{Corollary~#1}
\frefformat{vario}{\fancyrefdeflabelprefix}{Definition~#1}
\frefformat{vario}{\fancyrefobslabelprefix}{Observation~#1}
\frefformat{vario}{\fancyrefasmlabelprefix}{Assumption~#1}
\frefformat{vario}{\fancyrefasmslabelprefix}{Assumptions~#1}
\frefformat{vario}{\fancyreffiglabelprefix}{Figure~#1}
\frefformat{vario}{\fancyrefapplabelprefix}{Appendix~#1} 
\frefformat{vario}{\fancyrefproplabelprefix}{Proposition~#1}
\frefformat{vario}{\fancyrefalglabelprefix}{Algorithm~#1}
\frefformat{vario}{\fancyrefeqlabelprefix}{(#1)}
\frefformat{vario}{\fancyreftbllabelprefix}{Table~#1}
\frefformat{vario}{\fancyreftremabelprefix}{Remark~#1}

\safemath{\dictab}{[\,\dicta\,\,\dictb\,]}

\safemath{\ysig}{\bmy}
\safemath{\ysighat}{\hat{\ysig}}
\safemath{\ysigdim}{M}
\safemath{\xsig}{\bmx}
\safemath{\xsigdim}{N}
\safemath{\nx}{n_x}
\safemath{\zsig}{\bmz}
\safemath{\zsigdim}{\ysigdim}
\safemath{\rsig}{\bmr}
\safemath{\Adict}{\bA}
\safemath{\Adicttilde}{\widetilde{\Adict}}
\safemath{\Adictdim}{\outputdim\times\xsigdim}
\safemath{\avec}{\bma}
\safemath{\avectilde}{\tilde{\avec}}
\safemath{\Bdict}{\bB}
\safemath{\Bdicttilde}{\widetilde{\Bdict}}
\safemath{\Cdict}{\bC}
\safemath{\cvec}{\bmc}
\safemath{\Ddict}{\bD}
\safemath{\Ddictdim}{\ysigdim\times\xsigdim}
\safemath{\dvec}{\bmd}
\safemath{\Ddicttilde}{\widetilde{\bD}}
\safemath{\Bonb}{\bB}
\safemath{\bvec}{\bmb}
\safemath{\Bonbdim}{\ysigdim\times\ysigdim}
\safemath{\noise}{\bmn}
\safemath{\noisedim}{\ysigim}
\safemath{\err}{\bme}
\safemath{\errdim}{\ysigdim}
\safemath{\errset}{\setE}
\safemath{\nerr}{n_e}
\safemath{\delop}{\bP_\errset}
\safemath{\delopc}{\bP_{{\errset}^c}}

\safemath{\cplxi}{\imath}
\safemath{\cplxj}{\jmath}

\safemath{\dict}{\matD}
\safemath{\inputdim}{N}		
\safemath{\outputdim}{M}		
\safemath{\sparsity}{S}	
\safemath{\inputdimA}{{N_a}}	
\safemath{\inputdimB}{{N_b}}	
\safemath{\elemA}{{n_a}}	
\safemath{\elemB}{{n_b}}	
\safemath{\resA}{\matR_a}	
\safemath{\resB}{\matR_b}	
\safemath{\subD}{\matS} 
\safemath{\subA}{\matS_a} 
\safemath{\subB}{\matS_b} 
\safemath{\dicta}{\matA} 	
\safemath{\dictb}{\matB} 	
\safemath{\hollowS}{H}
\safemath{\hollowA}{H_a}
\safemath{\hollowB}{H_b}
\safemath{\cross}{Z}
\safemath{\coh}{\mu_d}			
\safemath{\coha}{\mu_a}			
\safemath{\cohb}{\mu_b}			
\safemath{\mubs}{\nu}	
\safemath{\cohm}{\mu_m} 
\safemath{\dictset}{\setD}	
\safemath{\dictsetp}{\dictset(\coh,\coha,\cohb)}	
\safemath{\dictsetgen}{\dictset_\text{gen}}
\safemath{\dictsetgenp}{\dictsetgen(\coh)}
\safemath{\dictsetonb}{\dictset_\text{onb}}
\safemath{\dictsetonbp}{\dictsetonb(\coh)}

\safemath{\leftside}{U}
\safemath{\rightsideA}{R_a}
\safemath{\rightsideB}{R_b}

\safemath{\indexS}{\setI_S} 

\safemath{\na}{n_a}			
\safemath{\nb}{n_b}			
\safemath{\coeffa}{p_i}	
\safemath{\coeffb}{q_j}	
\safemath{\seta}{\setP}		
\safemath{\setb}{\setQ}     
\safemath{\setw}{\setW}	
\safemath{\setz}{\setZ}	
\safemath{\cola}{\veca}		
\safemath{\colb}{\vecb}		
\safemath{\cold}{\vecd}		
\safemath{\inputvec}{\vecx} 	
\safemath{\error}{\vece}	
\safemath{\noiseout}{\vecz} 	
\safemath{\inputvecel}{x}
\safemath{\inputveca}{\vecx_a}
\safemath{\inputvecb}{\vecx_b}
\safemath{\outputvec}{\vecy}	
\safemath{\lambdamin}{\lambda_{\mathrm{min}}}

\safemath{\elltwo}{\ell_2}
\safemath{\ellone}{\ell_1}
\safemath{\ellzero}{\ell_0}
\safemath{\ellinf}{\ell_\infty}
\safemath{\ellinftilde}{\ell_{\widetilde\infty}}
\safemath{\licard}{Z(\coh,\coha,\cohb)}
\safemath{\xsol}{\hat{x}}
\safemath{\xbord}{x_b}		
\safemath{\xstat}{x_s}		
\safemath{\xstatLone}{\tilde{x}_s}
\safemath{\order}{\mathcal{O}} 
\safemath{\scales}{\Theta} 
\safemath{\ones}{\mathbf{1}} 
\safemath{\zeroes}{\mathbf{0}} 
\safemath{\thlone}{\kappa(\coh,\cohb)} 
\safemath{\constoneA}{\delta} 
\safemath{\constoneB}{\epsilon} 
\safemath{\nlarge}{L}				   
\safemath{\sumlarge}{S_\nlarge}
\safemath{\maxlarger}{P_\nlarge}	   
\safemath{\Pzero}{\textrm{P0}}	
\safemath{\Pone}{\textrm{P1}}
\safemath{\vecfir}{\vecw}			 
\safemath{\vecsec}{\vecz}
\safemath{\elvecfir}{w}              
\safemath{\elvecsec}{z}				 
\safemath{\nlargefir}{n}
\safemath{\normout}{\gamma}
\safemath{\auxfun}{h}
\safemath{\supp}{\textrm{supp}}

\safemath{\indexa}{\ell}
\safemath{\indexb}{r}
\safemath{\indexc}{i}
\safemath{\indexd}{j}

\safemath{\project}{P}

\newcommand{\mb}[1]{{\mathbf #1}}

\renewcommand{\bml}{\ensuremath{\boldsymbol \ell}}

\newcommand{\eg}[1]{\textcolor{green}{\bf[eg: #1]}}

\safemath{\LAMA}{\textrm{LAMA}}
\safemath{\MRT}{\textrm{MRT}}
\safemath{\betamax}{\beta^\text{max}_\setO}
\safemath{\betamaxno}{\beta^\text{max}}
\safemath{\betamin}{\beta^\text{min}_\setO}
\safemath{\betaminno}{\beta^\text{min}}

\safemath{\Nomin}{\No^\textnormal{min}(\beta)}
\safemath{\Nominnobeta}{\No^\text{min}}
\safemath{\Nomax}{\No^\textnormal{max}(\beta)}
\safemath{\Nomaxnobeta}{\No^\textnormal{max}}
\safemath{\EX}{E_\textnormal{x}}
\safemath{\EXP}{\EX^\textnormal{p}}
\safemath{\Eo}{E_0}

\safemath{\tmax}{{t_\textnormal{max}}}
\safemath{\MAP}{\textrm{MAP}}
\safemath{\IO}{\textrm{IO}}
\safemath{\JO}{\textrm{JO}}
\safemath{\Nopost}{N_{0}^\textnormal{post}}
\safemath{\MT}{U}
\safemath{\MR}{B}
\safemath{\Tran}{\textnormal{T}}
\safemath{\Herm}{\textnormal{H}}
\safemath{\row}{\textnormal{r}}
\safemath{\col}{\textnormal{c}}

\safemath{\NT}{N_\textnormal{T}}
\safemath{\DSNR}{\delta \textnormal{SNR}}
\safemath{\betaMOR}{\beta^{\star}}

\title{Semi-supervised t-{SNE} for Millimeter-wave Wireless Localization}

\author{
    \IEEEauthorblockN{
		Junquan Deng$^\text{1,*}$,
        Wei Shi$^\text{1}$,
        Jian Hu$^\text{1}$,
		Xianlong Jiao$^\text{2}$}\\
	
\IEEEauthorblockA{
		\small $^\text{1}$\textit{Sixty-third Research Institute, National University of Defense Technology, China, Emails:  {jqdeng@nudt.edu.cn}}\\
	    $^\text{2}$\textit{College of Computer Science, Chongqing University, China,  Email: {xljiao@cqu.edu.cn}}\\
		\thanks{This work  was supported in part by the National Science Foundation of China under grant 61901497 and 62072064, in part by Research Project of National University of Defense Technology under grant ZK 19-09.}
	}
}

\begin{document}
\maketitle
\begin{abstract}
We consider the mobile localization problem in future millimeter-wave wireless networks with distributed Base Stations~(BSs) based on multi-antenna channel state information~(CSI). For this problem, we propose a Semi-supervised t-distributed Stochastic Neighbor Embedding~(St-SNE) algorithm to directly embed the high-dimensional CSI samples into the 2D geographical map.  We evaluate the performance of St-SNE in a simulated urban outdoor millimeter-wave radio access network. Our results show that St-SNE achieves a mean localization error of 6.8\,m with only 5\% of labeled CSI samples in a 200$\times$200 m$^2$ area with a ray-tracing channel model. St-SNE does not require accurate synchronization among multiple BSs, and is promising for future large-scale millimeter-wave localization.
\end{abstract}


\section{Introduction}
The mobile location information of User Equipments~(UE) is critical for many smart city applications, including traffic monitoring, asset tracking, autonomous driving, emergency rescue and so forth. Currently, mobile localization heavily depends on global navigation satellite system~(GNSS) technologies, but GNSSs are not omnipotent for all kinds of applications and all scenarios. For example, a GNSS may fail to provide reliable position estimates due to signal blockage in dense urban areas.
Furthermore, continuous reception and detection of GNSS signals is a major part of battery consumption for many mobile devices. In some applications where a central entity needs to collect location information from massive user devices, GNSS is not adequate as users may refuse to report their GNSS information.
Due to these problems, we need to resort to other complementary methods for localization and positioning where GNSSs are not suitable.

Future 5G and beyond radio access networks~(RANs) are envisioned to be densely deployed with massive multiple-antenna Base Stations~(BSs) and high-frequency carrier frequencies, such as millimeter-wave~(mmWave), to provide ubiquitous ultra-fast and reliable wireless connections.
In addition to communication, the large-scale dense RAN infrastructure, the massive antennas and the wide high-frequency bands can also be leveraged for sensing and localization purposes~\cite{Mendrzik_SPAWC2019,Laoudias_Survey_2018,Wymeersch_WC_2017,Zhang_AI_5G}.

However, current 5G New Radio positioning techniques~\cite{Keating_Overview_2019} are based on triangulation with Angle-of-Arrival~(AOA) or trilateration with Time-Difference-of-Arrival~(TDOA) measurements at multiple BSs, which require accurate calibration and synchronization among RAN elements, and thus incur high deployment and maintenance costs for the operators. Moreover, the performance of these methods degrades in complex multi-path and obstructed environments.
Fingerprinting techniques~\cite{Chapre_2014,Sun_TVT_2018,Meng_LANMAN_2020} could be applied for cellular positioning in such challenging propagation environments. However, they require a large amount of dedicated and densely-sampled labeled measurements, so scale poorly to large areas and render automatic operation and maintenance in dynamic environments challenging.
%
As unlabeled samples are easier to be collected, semi-supervised learning methods~\cite{zhu05survey} with both labeled and unlabeled data are promising for automatic large-scale localization. Semi-supervised learning has attracted increasing attention for positioning based on Received Signal Strength Indicator~(RSSI) fingerprints\cite{Pan_TPAMI_2012,Pulkkinen_2011}. However, research work on utilizing multi-antenna Channel State Information~(CSI)~\cite{Bast_VTC2020,WSA_mMIMO_data_2020} for high-precision and seamless localization in a networking environment is still limited.

Recently, channel charting~\cite{CC_2017,MPCC_Deng_2018} has been proposed to use sporadically collected multi-antenna CSI samples from unknown locations to construct a channel chart that can provide relative position information among UEs.
Channel charting exploits the fact that high-dimensional multi-antenna CSI strongly depends on the low-dimensional 2D/3D UE location as a result of physical law of radio propagation.
Several works~\cite{Studer_SPAWC2019, Studer_Allerton2019, SSMPCC2021} have dedicated to equip channel charting with absolute positioning capabilities.
In~\cite{Studer_SPAWC2019}, a semi-supervised auto-encoder has been proposed to utilize a subset CSI of samples with known spatial locations, as well as mobility side information to locate UEs in a single cell.
In~\cite{Studer_Allerton2019}, Sammon's mapping~(SM) and Siamese neural network are combined in an unified channel charting framework to provide both unsupervised relative localization, and semi-supervised absolute positioning.
Unfortunately, in a scenario with challenging non-Line-of-Sight~(NLOS) propagation channels, the localization performance of~\cite{Studer_SPAWC2019,Studer_Allerton2019} is not satisfying. In addition, these two methods have not considered the more realistic large-scale multi-cell scenario with multiple BSs.
For multi-cell localization, a semi-supervised multi-point channel charting~(SS-MPCC) framework has been put forward in~\cite{SSMPCC2021}. In SS-MPCC, multiple distributed BSs collect wireless data from mobile UEs and learn an aligned channel chart in which historical and real-time mobile locations can be determined. It uses Semi-supervised Laplacian Eigenmap~(SLE) with both labeled CSI and time-stamp information to increase the smoothness and trustworthiness of the learned channel chart, and greatly improves the localization performance.

In this paper, we propose a new machine learning algorithm, called Semi-supervised t-distributed Stochastic Neighbor Embedding~(St-SNE for short), for 5G and beyond mmWave localization based on the SS-MPCC framework.
The original t-distributed Stochastic Neighbor Embedding~(t-SNE) technique is a statistical method for visualizing high-dimensional data, such as images, audios and DNA sequences, by giving each sample a location in a two or three-dimensional~(2D/3D) map, as long as there is a suitable similarity metric for two samples.
It has been used in a wide range of applications, including  natural language processing, image pattern analysis, bioinformatics, etc.  Directly applying t-SNE to map the CSI data to a 2D map can also provide some interesting results as in~\cite{MPCC_Deng_2018}, but the generated 2D map cannot be used for practical positioning applications. To this end, in St-SNE, we first consider using a few position labels to guide the learning process of t-SNE to produce a map that relate to the true geographical UE positions.
To demonstrate the effectiveness of St-SNE, we perform simulations in an urban outdoor multi-cell mmWave network, using a simulator that models UE distribution in a Manhattan street grid and the complex radio propagation conditions. We show that St-SNE significantly improves the localization performance compared with other conventional techniques.

\section{MmWave Localization Framework}

\begin{figure}[tp]
	\centering
	\includegraphics[width=0.425\textwidth]{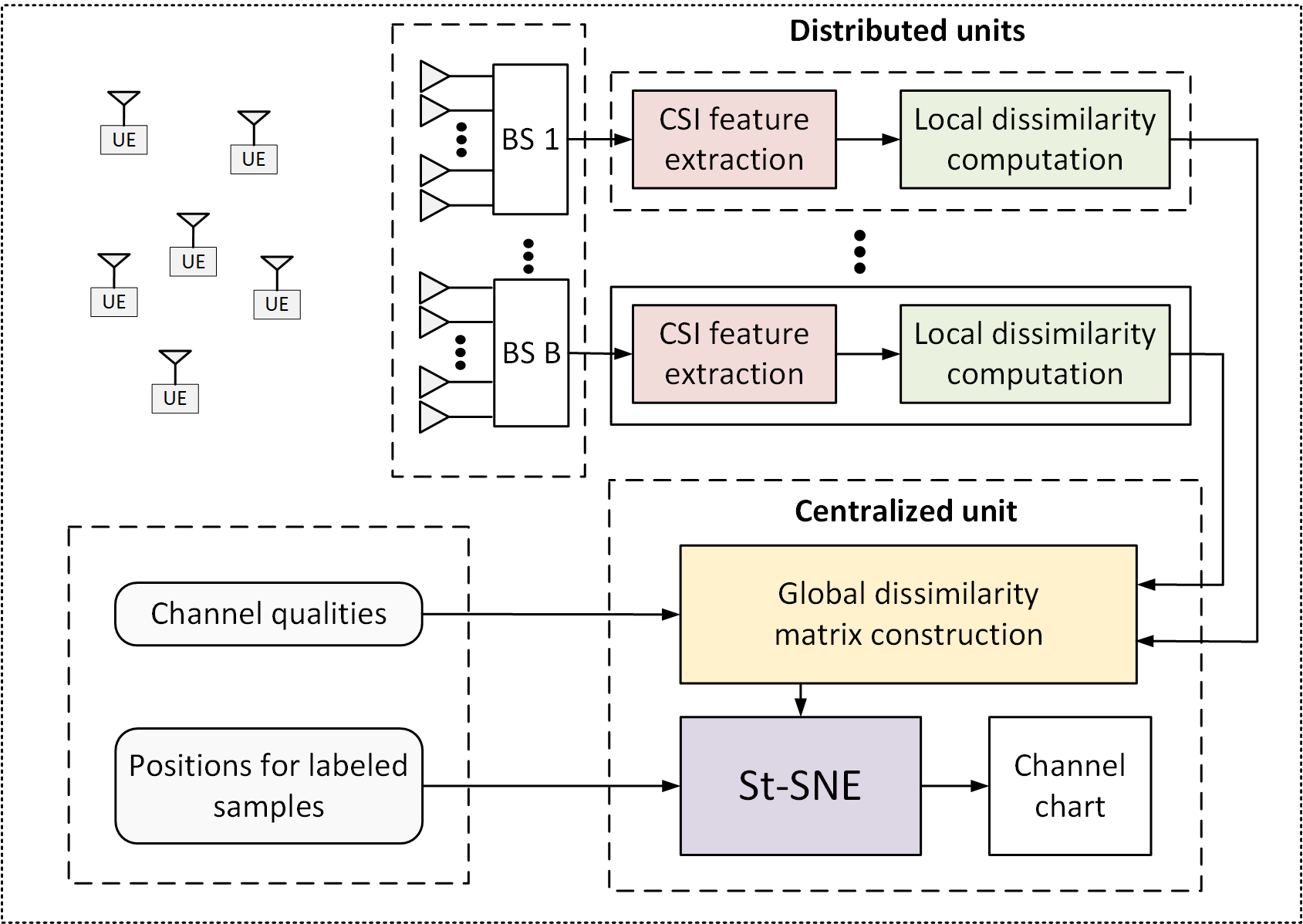}
	\caption{Semi-supervised multi-point channel charting framework for St-SNE.} 
	\label{fig:framework}
\end{figure}

The multi-cell mmWave localization framework is shown in Fig.~\ref{fig:framework}. Similar to~\cite{SSMPCC2021}, it consists of a labeled and unlabeled CSI samples collection procedure, feature extraction and local dissimilarity matrix construction procedures at distributed units~(DUs), as well as global dissimilarity matrix construction and semi-supervised manifold learning procedures conducted at the centralized unit~(CU).

\subsection{Channel Model}

We consider a typical 5G mmWave cellular network where multiple UEs move in the coverage area of multiple BSs. The UEs send Orthogonal Frequency Division Multiplexing~(OFDM) pilot signals to BSs for channel estimation. An estimated multi-antenna channel vector at time $t$ on a sub-carrier with frequency $f$ is modelled as
\begin{equation}\label{eq:channelmodel}
\mb h_{t,f} = \sum\nolimits_{k=1}^{K}{ {\alpha}^{(k)}_{t,f}  \mathbf{s}(\phi^{(k)}_t) } + \mb e,
\end{equation}
where $K$ denotes the number of multi-path components, $\phi^{(k)}_t$ the
direction-of-arrival (DOA) of the $k$th propagation path, and ${\alpha}^{(k)}_{t,f}$ a random complex gain for the $k$th path.
In addition, $\mb e$ represents the channel estimation error, and $\mathbf{s}(\phi)$  the array steering vector~(ASV). We assume an $M$-antenna uniform linear array is used at a BS, and the ASV is
\begin{equation}\label{eq:aBS}
\begin{aligned}
\mb s(\phi)\!  &=\!\big[1, {e}^{j \frac{2\pi}{\lambda}s\sin(\phi)},\ldots,{e}^{j(\!M-1\!)\frac{2\pi}{\lambda}s\sin(\phi)} \big]^{\rm T},
\end{aligned}
\end{equation}
with $\lambda$ the carrier wavelength and $s$ the antenna spacing.
\subsection{CSI Collection and Dissimilarity Matrix Construction}
The channel vector $\mb h_{t,f} $ itself changes rapidly when the UE moves
across multiple wavelengths. In comparison, the multipath DOAs $\{\phi^{(l)}_{t,f}\}$ and  gains $\{{\alpha}^{(l)}_t\}$ change more slowly. We use the frequency-domain covariance
$
\mb C = \mathbb{E}_f \big[\mb h_{t,f} {\mb h^{\rm H}_{t,f}}\big] 
$
as the CSI sample for localization, as it depends on the multipath DOAs and gains, and changes slowly when UE moves.

We assume there are $B$ BSs, they will collect CSI samples from multiple UEs during network operation. There are two types of CSI samples, with or without position labels. The first type has the corresponding UE position information which can be gathered in a dedicated site survey or reported by UEs with GNSS capability, while the second type is collected from UEs which need to be located or tracked.
After data collection, we assume there are $L$ labeled CSI samples at BS $b$, denoted by $\mathcal{L}^{(b)} =\{{\mathbf{C}}^{(b)}_{1},\ldots,{\mathbf{C}}^{(b)}_{L}\}$. The corresponding ground-truth location coordinates are $\mb Y = [\mb y_1,\ldots,\mb y_L]$.
The in total $U$ unlabeled samples collected at BS $b$ can be denoted as $\mathcal{U}^{(b)} =\{\mathbf{C}^{(b)}_{L+1},\ldots,\mathbf{C}^{(b)}_{L+U}\}$.
The corresponding unknown locations for the unlabeled samples are denoted by $\hat{\mb Y} = [\mb y_{L+1},\ldots,\mb y_{L+U}]$. We assume that multiple BSs can identity the CSI samples coming from a specific UE at a same time interval via UE ID and loose time synchronization. If an UE's pilot signals are not received by BS $b$ at a time, the corresponding CSI sample ${\mathbf{C}}^{(b)}_{i}$ is set to zeros. Each BS will utilize a feature extraction function to extract the power angular profile (PAP) of multipath components~(MPCs) hidden in the CSI sample and compute the dissimilarities among its collected samples. The CU then fuse the local dissimilarities and channel qualities information reported by BSs to construct a global consistent dissimilarity matrix $\mb D \in \mathbb{R}^{N\times N}$, with $N = L+U$ . The details of the feature extraction, dissimilarity metric and fusion procedures can found in~\cite{SSMPCC2021}. The global dissimilarity matrix $\mb D$ characterizes how similar the uplink radio channel conditions for different UE positions.

%

\section{Semi-Supervised t-SNE}
\label{sec:St-SNE}

The {\emph{t}-SNE}~\cite{tSNE} is an effective dimensional reduction
and manifold learning algorithm widely used for visualizing
high-dimensional data. This method learns a low-dimensional representation $\mb Z = [\mb z_1, \ldots, \mb z_N]$ of
the $N$ data points by minimizing the divergence between two distributions,
i.e., a distribution that characterizes pairwise similarities of the
input data points, and a distribution that characterizes pairwise
similarities of the corresponding low-dimensional points in the
representation space. To this end, {\emph{t}-SNE} defines a symmetric
probability matrix $\mb P$ with elements $\{p_{nm}\}_{n,m=1}^N$ that measures the
pairwise similarity between $n$th and $m$th input data points. Denote $\{d_{nm}\}_{n,m=1}^N$ the entities of the dissimilarity matrix $\mb D$, for $n
\neq m$, we have
\begin{equation}
\begin{aligned}
p_{nm} & = \frac{1}{2}(p_{m|n} + p_{n|m}) \\
       & = \frac{1}{2}\!\!\left(\!
\frac{{\rm e}^{-d_{mn}^2/2\sigma_n^2}}{\sum\nolimits_{k\neq n} {\rm e}^{-d_{kn}^2/2\sigma_n^2}} \! +\!
\frac{{\rm e}^{-d_{nm}^2/2\sigma_m^2}}{\sum\nolimits_{k\neq m} {\rm e}^{-d_{km}^2/2\sigma_m^2}}
\!\right),
\end{aligned}\label{eq3}
\end{equation}
and $p_{nm} = 0$ for $n = m$, with $p_{m|n}$ the conditional probability the $m$th point is a neighbor of the $n$th point .
The bandwidth of the Gaussian kernels ${\sigma_n}$ for the $n$th data
point is set so that $2^{-\sum\nolimits_m p_{m|n} \log_2 p_{m|n}}$ equals to a given
parameter $k_{t}$ called \emph{perplexity}, e.g., using a 1D search method. The perplexity can
be interpreted as a measure of the effective number of neighbors taken
into account.

In the representation space, a matrix $\mb Q$ with
element $\{q_{nm}\}_{n,m=1}^N$ which measures the similarity between
$\mb z_n$ and $\mb z_m$, is computed using a normalized
\emph{t}-distribution kernel by
\begin{equation}
\begin{aligned}
q_{nm} = \frac{(1+\|\mb z_n - \mb z_m\|^2_2)^{-1}}{ \sum\nolimits_{l \neq k} (1+\|\mb z_l - \mb z_k\|^2_2)^{-1} }.
\end{aligned}\label{eq4}
\end{equation}
The objective of \emph{t}-SNE is to find a representation by minimizing the Kullback-Leibler~(KL)
divergence between the two distributions $\mb P$ and $\mb Q$. The optimization problem is
\begin{align*}
(\text{P1})
\left\{\begin{array}{ll}
\underset{\mb Z }{\text{minimize}} & f_{ t\text{-SNE}}(\mb Z) = \sum_n \! \sum_m p_{nm} \! \log\frac{p_{nm}}{q_{nm}} ,\\[0.05cm]
\text{subject to} &  \sum_{n=1}^{N} \mb z_n = \mb 0.
\end{array}\right.
\end{align*}
The objective function $f_{ t\text{-SNE}}(\mb Z)$ can be minimized by
gradient descent. Note that the KL divergence is not convex. Different
initializations will possibly end up in different local minima of
$f_{\rm tSNE}(\mb Z)$. Hence, it is useful to try different seeds and
choose the result with the lowest KL divergence. {\emph{t}-SNE} is
computationally expensive, especially for large-scale data sets.
Larger perplexities lead to more neighbors and less sensitivity to small-scale
structure. By contrast, a lower perplexity focuses on a smaller number
of neighbors, and thus ignores more global information favouring the
local neighborhood preservation. For a larger data set, larger
perplexities are required.

Directly solve problem~P1 can not provide a 2D map with estimated UE position information.
To equip t-SNE with localization capability, we need to use some position labels to govern the learning process of t-SNE.
To this end, we formalize the following semi-supervised t-SNE problem,
\begin{align*}
(\text{P2})
\left\{\begin{array}{ll}
\underset{\mb Z }{\text{minimize}} & f_{ t\text{-SNE}}(\mb Z) = \sum_n \! \sum_m p_{nm} \! \log\frac{p_{nm}}{q_{nm}} ,\\[0.05cm]
\text{subject to} & \mb z_i = \mb y_i, i \in {\cal{L}}, {\cal{L}} = \{1,\ldots,L\}.
\end{array}\right.
\end{align*}
As the labeled CSI samples are restricted to be mapped to their corresponding true UE locations, minimizing the cost function would probably lead to an UE position map, and
the unknown locations of unlabeled samples can be estimated.
As shown in~\cite{tSNE}, the gradient of the Kullback-Leibler divergence between $\mb P$ and $\mb Q$  is given by
\begin{equation}\label{eq5}
  \frac{ \mathrm{d} f_{ t\text{-SNE}}(\mb Z)}{\mathrm{d} \mb z_n} = 4 \sum_m \frac{(p_{nm}-q_{nm})(\mb z_n - \mb z_m)}{(1+||\mb z_n - \mb z_m||^2)}.
\end{equation}
To solve problem~P2, we have devised a gradient descent algorithm as summarized in Algorithm~\ref{alg1}. In Algorithm~\ref{alg1}, the learning process is controlled by four parameters, the perplexity $k_t$, iteration number $T$, learning rate $\eta$ and momentum $\alpha$. They will be investigated in Section \ref{sec:para}. At the end of each iteration, the coordinates of the labeled samples in the 2D map are forced to be equal to their position labels, so the learned 2D map would be aligned in the geographical space. The computation complexity of one iteration in Algorithm~\ref{alg1} is ${\cal{O}}(N^2)$~\cite{tSNE}, and the overall complexity is ${\cal{O}}(TN^2)$.

\begin{algorithm}[t]
	\begin{algorithmic}[1]
		\State \textbf{Inputs:} $\mb D \in \mathbb{R}^{N\times N}$, $\{\mb y_1,\ldots,\mb y_L\}$, ${\cal{L}} = \{1,\ldots,L\}$
        \State \textbf{Cost function parameter}: $k_{t}$
        \State \textbf{Optimization parameters}: $T$, $\eta$ and $\alpha$
		\State \textbf{Initialize:} $\mb Z^{(0)} = [\mb z^{(0)}_1, \ldots, \mb z^{(0)}_N]$, $\mb Z^{(-1)} = \mb Z^{(0)}$, with \\
        \quad \, $\mb z^{(0)}_i = \mb y_i$, for $ i \in {\cal{L}}$, \\
        \quad \, $\mb z^{(0)}_i = \mb y_n$, for $i \notin {\cal{L}}$ and $d_{n,i}$ is smallest for all $n\neq i$
		\State Use binary search to determine the kernel sizes \{$\delta_n$\}$_{n=1}^{N}$, and compute the probability matrix $\mb P$ using ~\eqref{eq3}																										
		\For {$t = 1,\ldots, T$}																										
		\State Compute probability matrix $\mb Q$ using ~\eqref{eq4}		
        \State Compute	$ \nabla = [\frac{ \mathrm{d} f_{ t\text{-SNE}}(\mb Z)}{\mathrm{d} \mb z_1} , \ldots, \frac{ \mathrm{d} f_{ t\text{-SNE}}(\mb Z)}{\mathrm{d} \mb z_N}  ] $ using ~\eqref{eq5}		
        \State Update $\mb Z^{(t)} = \mb Z^{(t-1)} + \eta \nabla + \alpha (\mb Z^{(t-1)}  - \mb Z^{(t-2)} )$
        \State Set $\mb z^{(t)}_i = \mb y_i$, for $ i \in {\cal{L}}$	
		\EndFor
		\State \textbf{return:} $\mb Z^{(T)}$		
	\end{algorithmic}
	\caption{The semi-supervised t-SNE algorithm}
	\label{alg1}
\end{algorithm}

\section{Simulation Results}\label{sec:sim}
We now demonstrate the efficacy of the proposed St-SNE algorithm for CSI-based localization with a subset of CSI samples with marked positions, investigate the effects of its learning parameters on the performance, and compare it with some other typical CSI-based localization methods.
\subsection{Simulated Scenario and Evaluation Metric}

\begin{figure}[tp]
	\centering
	\includegraphics[width=0.35\textwidth]{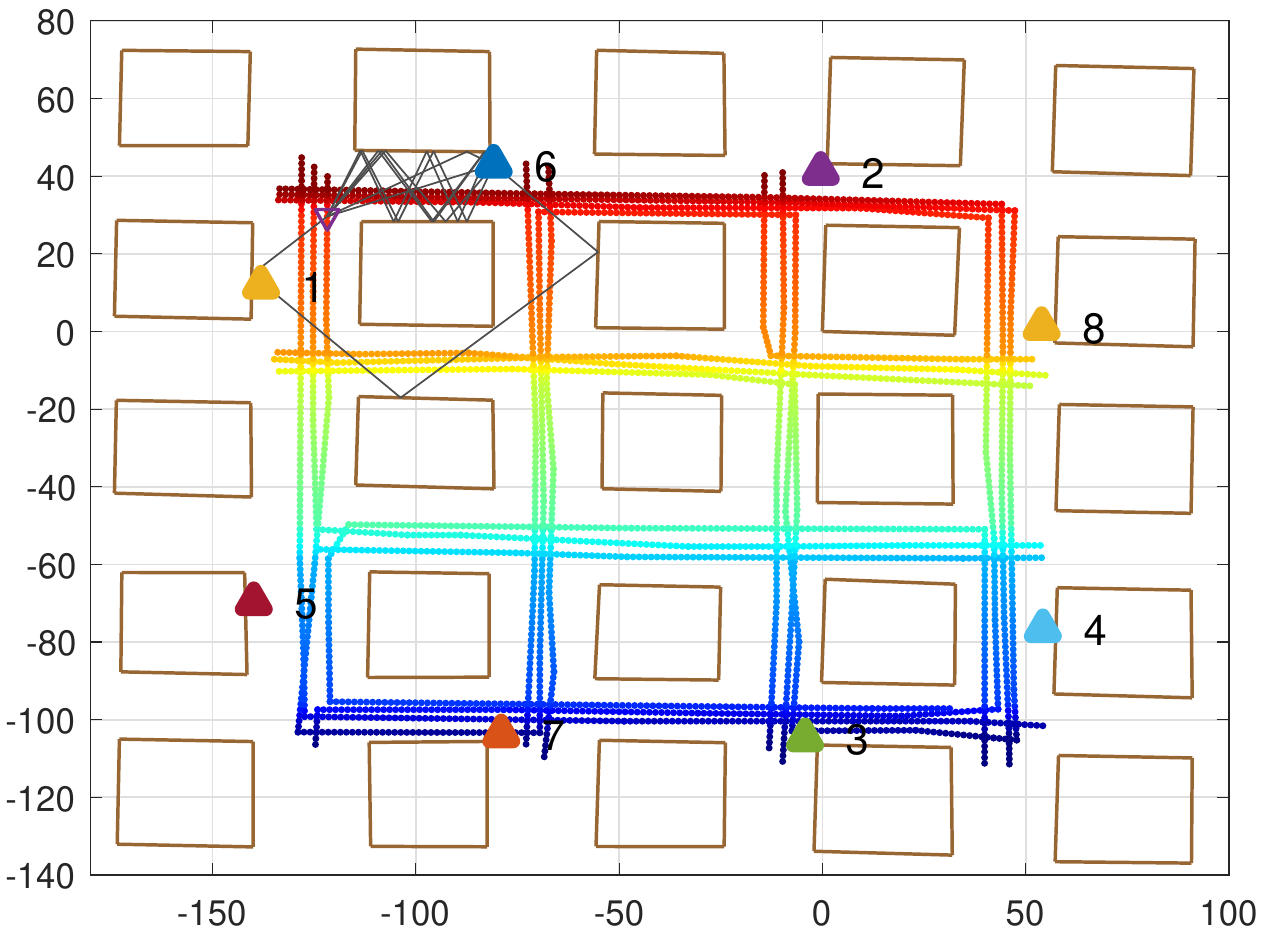}
	\caption{Distribution of 8 BSs marked by numbers, the traces of UEs during CSI sample collection represented by dots,
and multipath components for the channel between 6th BS and a UE location via ray-tracing.
The mobile locations are color-mapped for visualization purpose. }
	\label{fig:scenario}
\end{figure}

We consider a dense urban outdoor multi-cell mmWave network scenario as depicted in Fig.~\ref{fig:scenario}. BSs are below rooftop, and signals will be reflected or blocked by the walls. A ray-tracing channel model is used to generate the multi-path channels.
The reflection coefficients are computed based on the Fresnel equation and reflections with up to five bounces are taken into account. The relative permittivities of building walls are uniformly distributed between 3 and 7. There are 8 mMIMO BSs equipped with ULAs, each has $16$ elements with half-wavelength spacing. The antenna arrays of the BSs are oriented perpendicular to the building surfaces where the BSs are mounted. We collect CSI samples from $U=1425$ unknown positions among UE traces and the average distance between neighboring sampled locations is approximately 2~m. A covariance matrix is estimated over $32$ realizations of the channel vector $\mb h$ over multiple adjacent subcarriers.  For semi-supervised learning, additional $L=75$ samples are generated randomly on the roads. More simulation parameters are listed in Table~\ref{tbl:scenario}.
To measure the performance of CSI-based localization methods, we use the mean localization error (MLE) metric, which is
$
\textit{MLE} = \frac{1}{U}\sum_{n=1}^U \|\mb z_{L+n}-\mb y_{L+n}\|_2.
$

\begin{table}[t]
	\centering
	\renewcommand{\arraystretch}{1.2}
	\caption{Basic simulation Parameters}
	\begin{tabular}{@{}lc|clcc@{}}
		\toprule
		{Parameter}        & {Value}      & {Parameter}     & {Value}  \\
		\midrule
		Carrier frequency    &  28 GHz    &UE pilot Tx power        & 23 dBm       \\ 
		System bandwidth     &  256 MHz   &UE antenna pattern & Omnidirectional   \\ 
		Subcarrier number    &  128       &BS antenna pattern & Cosine response \\
		\bottomrule
	\end{tabular}
	\label{tbl:scenario}
\end{table}

\subsection{Algorithm Parameters} \label{sec:para}
\begin{figure}[tp]
\centering
\includegraphics[width=0.8\linewidth]{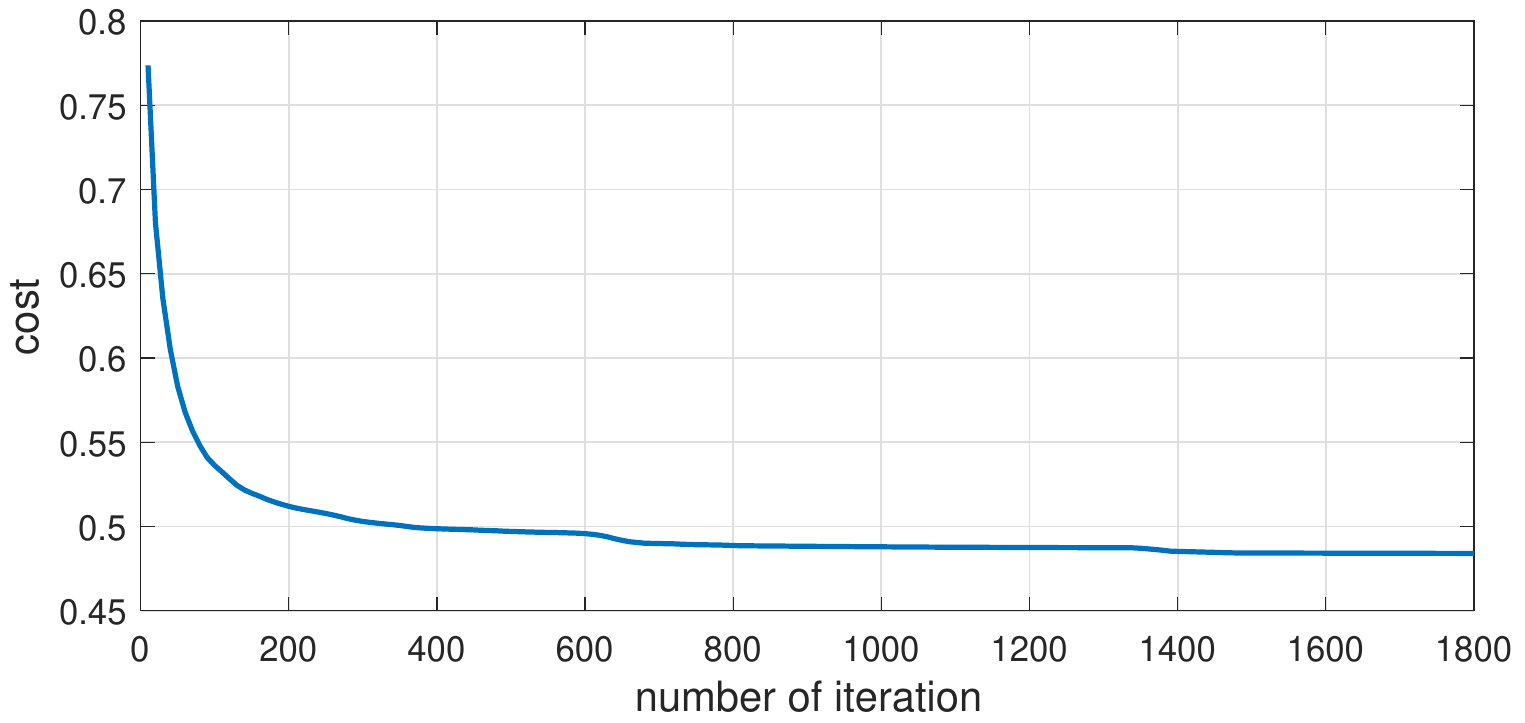}
\caption{St-SNE iteration process with perplexity $k_t = 30$, learning rate $\eta = 1000$, and momentum $\alpha = 0.6$.}\label{fig:Iter}
\end{figure}

\begin{figure}[tp]
\centering
\includegraphics[width=0.8\linewidth]{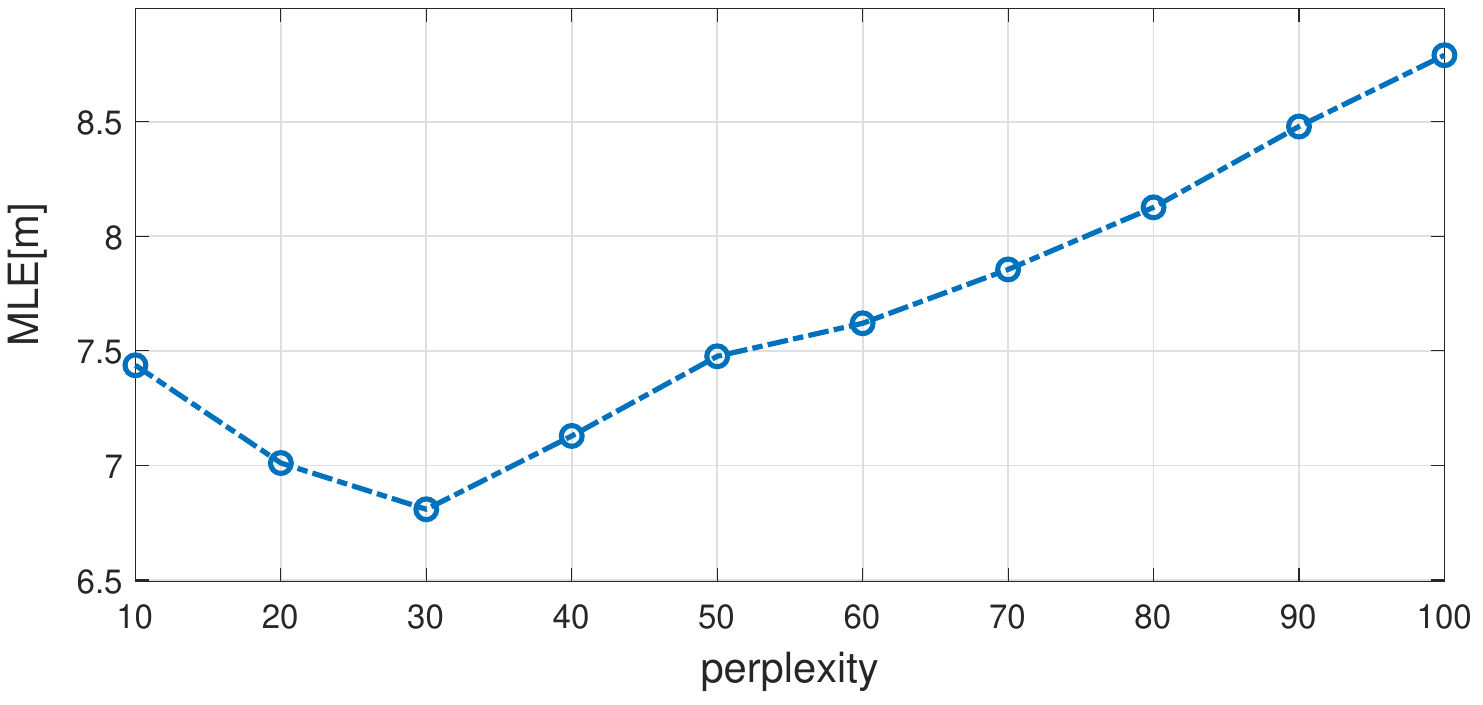}
\caption{The effects of perplexity $k_t$ in St-SNE.}\label{fig:kt}
\end{figure}

First, with a fixed perplexity $k_t$, we run the St-SNE algorithm with different learning rate $\eta$ and momentum $\alpha$. We found that a small $\eta$ will take more iterations for St-SNE to converge, while a large $\eta$ cannot find a local minimum for the cost function. It turns out that $\eta = 1000$ is a suitable option after trial and error. We use a typical momentum $\alpha = 0.6$ as in~\cite{tSNE} to accelerate the optimization process. Fig.~\ref{fig:Iter} shows that with such a setting, the algorithm converged after about 1500 iterations. So we set $T = 2000$ the max iteration number. We then change the perplexity $k_t$ and run the algorithm 10 times. The MLE vs perplexity curve is shown in Fig.~\ref{fig:kt}. A small perplexity will lead to the so called \emph{Crowding} problem~\cite{tSNE}, while a large one cannot reveal the manifold details of the 2D map. As shown in Fig.~\ref{fig:kt}, $k_t = 30$ can preserve both the local manifold details and the global structure, and leads to a minimum MLE of 6.8~m.

\subsection{Performance Comparison}

\begin{figure*}[tp]
	\centering
	\begin{subfigure}{0.3\textwidth}
		\centering
		\includegraphics[width=0.95\linewidth]{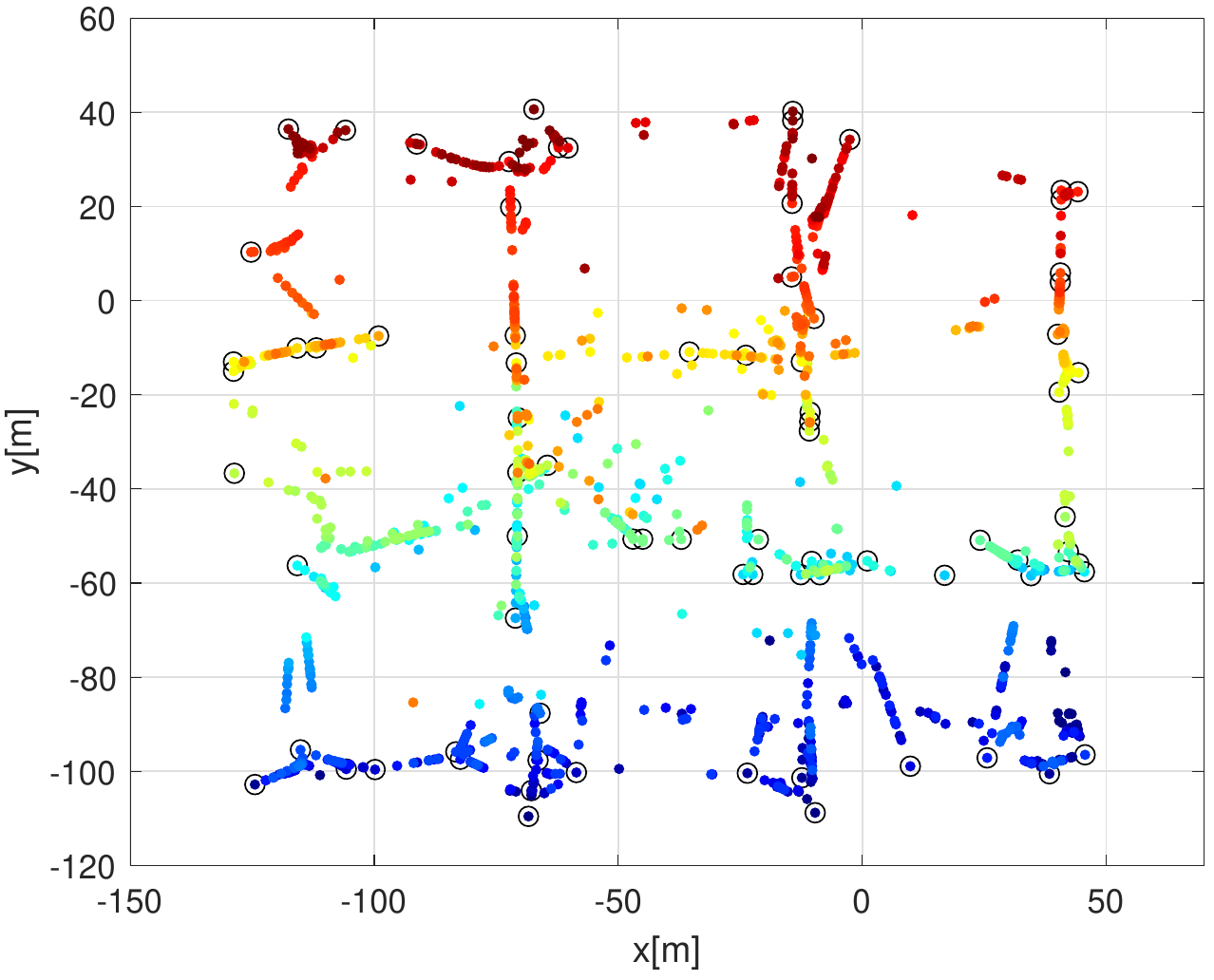}
		\caption{kNN}
	\end{subfigure}
	\hfill
	\begin{subfigure}{0.3\textwidth}
		\centering
		\includegraphics[width=0.95\linewidth]{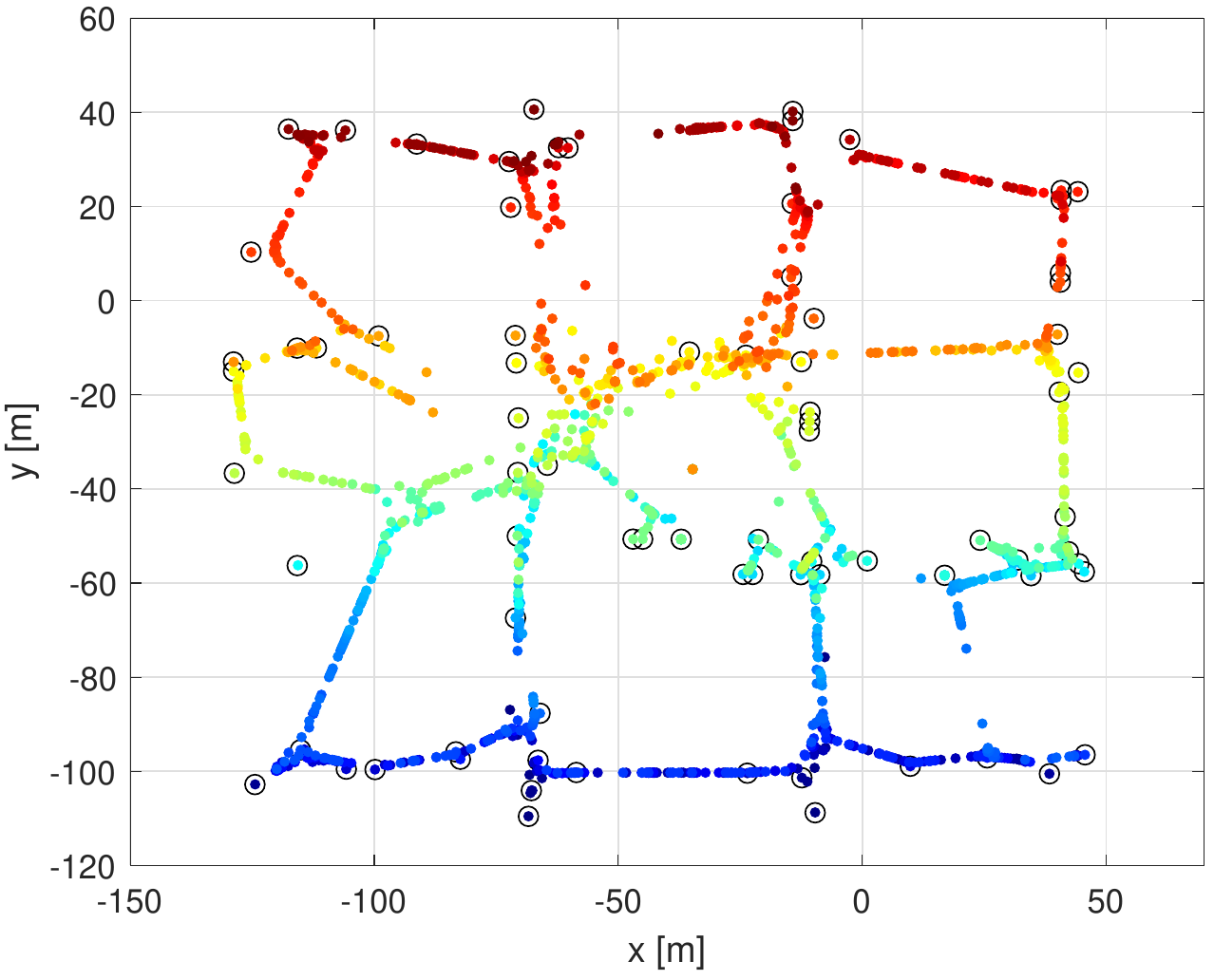}
		\caption{SLE}
	\end{subfigure}
	\hfill
	\begin{subfigure}{0.3\textwidth}
		\centering
		\includegraphics[width=0.95\textwidth]{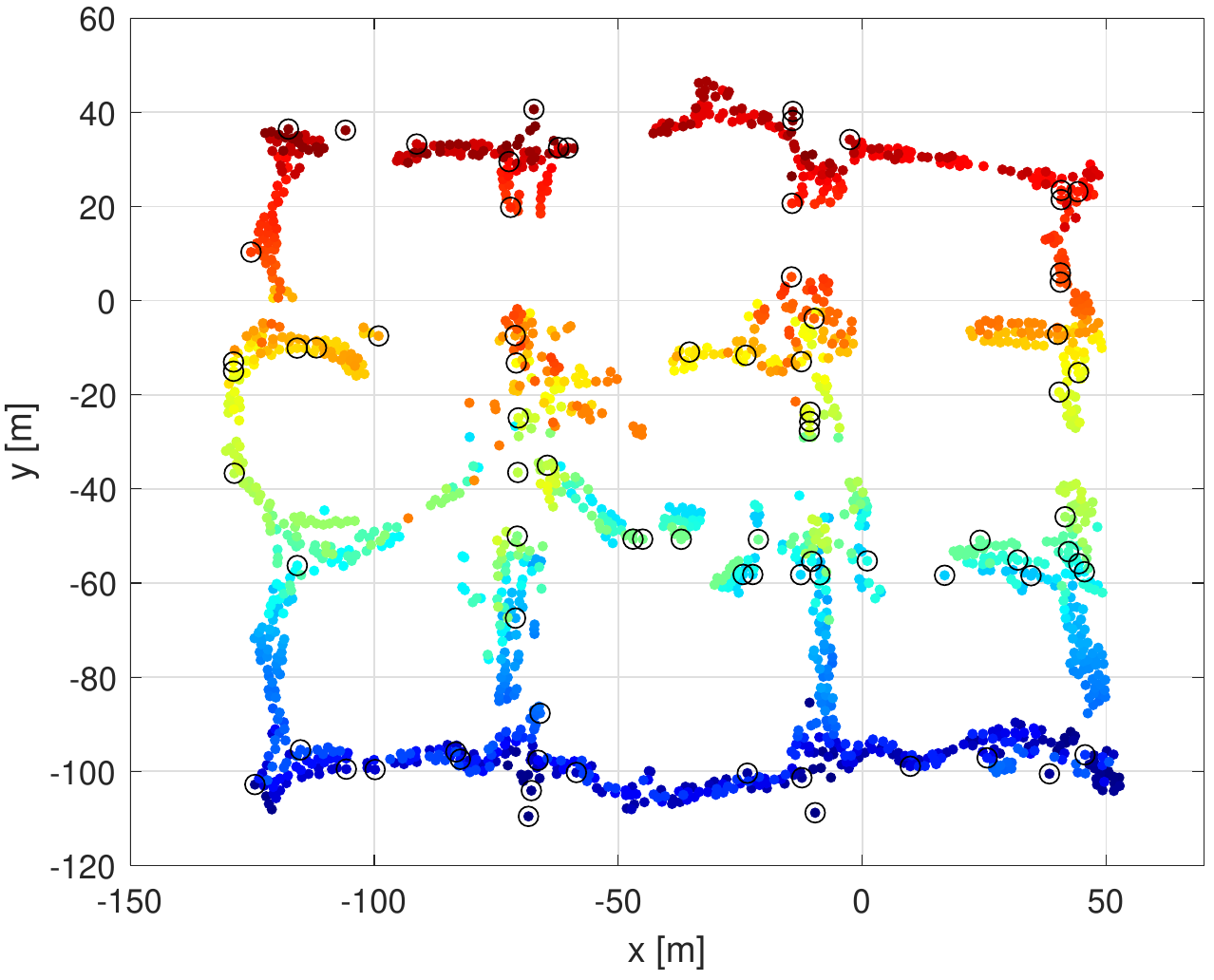}
		\caption{St-SNE}
	\end{subfigure}
	\caption{Visualization of the localization performance with different semi-supervised machine learning methods. The points with black circles represent the labeled UE positions, while other points represent the estimated positions for unlabeled samples. The MLEs are 10.9~m, 10.5~m and 6.8~m for (a) kNN, (b) SLE and (c) St-SNE respectively.
	}
	\label{fig:CC}
	\vspace{-0.2cm}
\end{figure*}

We now investigate the performance of the proposed St-SNE method for mmWave localization in more detail, and compare it with kNN~\cite{KNN_ICC2020} and semi-supervised Laplacian Eigenmap~(SLE)~\cite{SSMPCC2021} method. kNN is widely used in RSSI-based and CSI-based fingerprinting. Though kNN is quite simple, recent research shows that it demonstrated the best performance in localization accuracy among a wide range of machine learning methods in a complex environment~\cite{Rashdan_2020}. SLE is shown to be better than kNN in~\cite{SSMPCC2021}. For kNN, a small $k$ can obtain good performance~\cite{KNN_ICC2020} and we use $k=3$ neighbors which achieves smallest MLE in kNN here.

\begin{figure}[tp]
\centering
\includegraphics[width=0.8\linewidth]{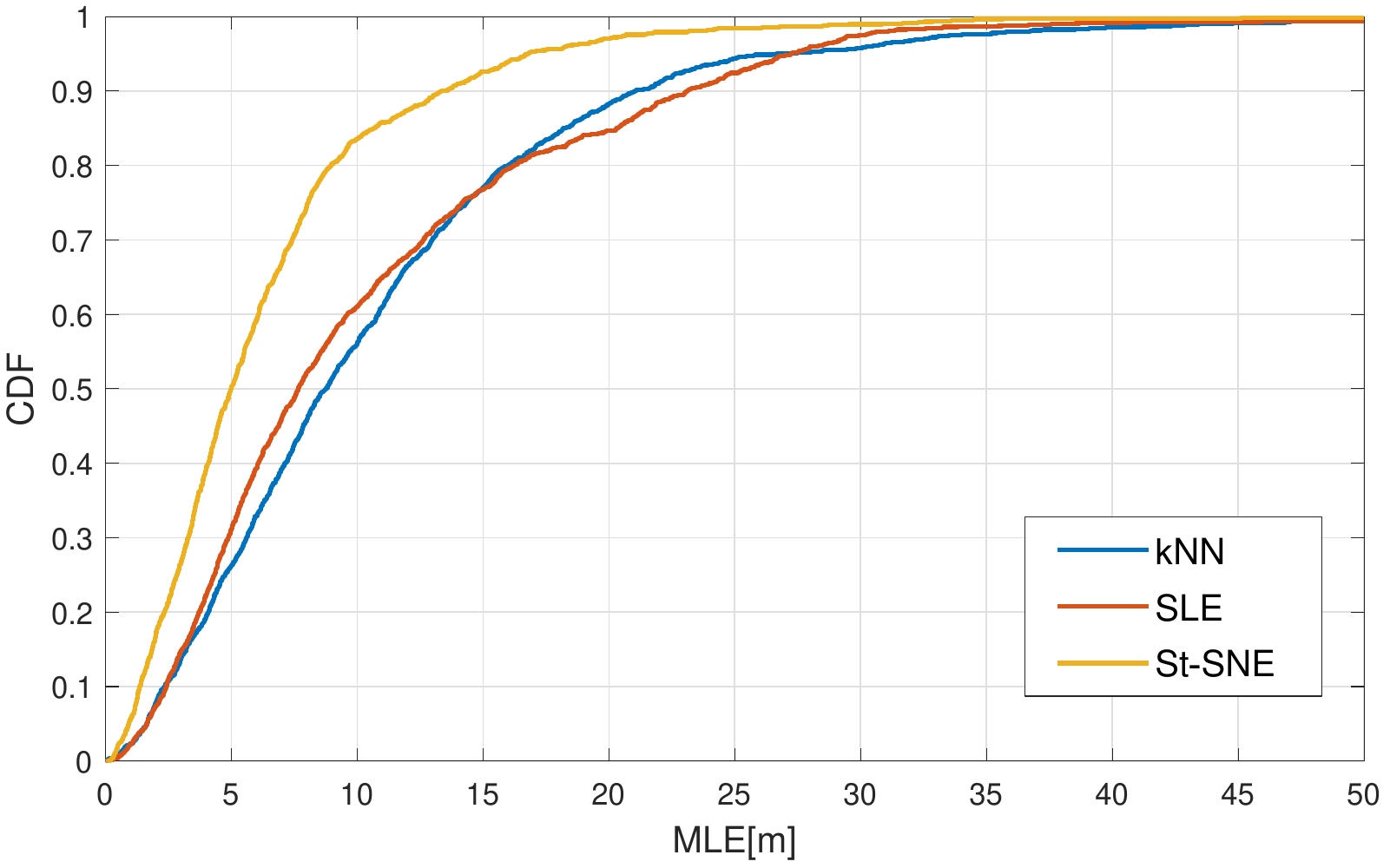}
\caption{Cumulative Distribution Function~(CDF) of localization errors.}\label{fig:CDF}
\end{figure}

The localization results with these three methods are illustrated in Fig.~\ref{fig:CC}, and the Cumulative Distribution Functions~(CDF) of localization errors are shown in Fig.~\ref{fig:CDF}. Compare the maps in Fig.~\ref{fig:CC} to the ground-truth map in Fig.~\ref{fig:scenario}, we see that positions of points far from the labeled anchors cannot be accurately estimated via kNN and SLE. It can be seen in Fig.~\ref{fig:CDF} that about 23\% of unlabeled points have a localization error larger than 15~m. Compared to kNN and SLE, St-SNE greatly reduces the errors of those points, with only 7\% of unlabeled points having a error larger than 15~m, leading to approximate 38\% and 35\% localization error reductions.

%
\section{Conclusions}
We have proposed a machine learning method St-SNE for mmWave multi-cell mobile localization. This method directly embeds the high-dimensional multi-antenna CSI samples into the 2D geographical map by governing the self-learning process of t-SNE with a few position labels. Through experiments in a simulated urban outdoor mmWave network, we have shown that St-SNE is able to perform accurate large-scale mmWave localization for scenarios with realistic UE distributions, even with a small portion of labeled data.
St-SNE is scalable and automatic in the sense that it could be implemented for multi-cell networks, with spatially sparse labeled samples, and does not require accurate network synchronization.
One drawback of St-SNE is its high computation complexity compared with kNN and SLE.  A prospective research direction would be to use a graph or tree method to accelerate its computation of the probability matrixes.

\bibliographystyle{IEEEtran}
\bibliography{ICCCS2022}

\begin{thebibliography}{10}
\providecommand{\url}[1]{#1}
\csname url@samestyle\endcsname
\providecommand{\newblock}{\relax}
\providecommand{\bibinfo}[2]{#2}
\providecommand{\BIBentrySTDinterwordspacing}{\spaceskip=0pt\relax}
\providecommand{\BIBentryALTinterwordstretchfactor}{4}
\providecommand{\BIBentryALTinterwordspacing}{\spaceskip=\fontdimen2\font plus
\BIBentryALTinterwordstretchfactor\fontdimen3\font minus
  \fontdimen4\font\relax}
\providecommand{\BIBforeignlanguage}[2]{{%
\expandafter\ifx\csname l@#1\endcsname\relax
\typeout{** WARNING: IEEEtran.bst: No hyphenation pattern has been}%
\typeout{** loaded for the language `#1'. Using the pattern for}%
\typeout{** the default language instead.}%
\else
\language=\csname l@#1\endcsname
\fi
#2}}
\providecommand{\BIBdecl}{\relax}
\BIBdecl

\bibitem{Mendrzik_SPAWC2019}
R.~{Mendrzik}, F.~{Meyer}, G.~{Bauch}, and M.~{Win}, ``Localization, mapping,
  and synchronization in {5G} millimeter wave massive {MIMO} systems,'' in
  \emph{IEEE 20th International Workshop on Signal Processing Advances in
  Wireless Communications (SPAWC)}, July 2019, pp. 1--5.

\bibitem{Laoudias_Survey_2018}
C.~{Laoudias}, A.~{Moreira}, S.~{Kim}, S.~{Lee}, L.~{Wirola}, and
  C.~{Fischione}, ``A survey of enabling technologies for network localization,
  tracking, and navigation,'' \emph{IEEE Communications Surveys Tutorials},
  vol.~20, no.~4, pp. 3607--3644, Fourthquarter 2018.

\bibitem{Wymeersch_WC_2017}
H.~{Wymeersch}, G.~{Seco-Granados}, G.~{Destino}, D.~{Dardari}, and
  F.~{Tufvesson}, ``{5G} {mmWave} positioning for vehicular networks,''
  \emph{IEEE Wireless Communications}, vol.~24, no.~6, pp. 80--86, Dec 2017.

\bibitem{Zhang_AI_5G}
C.~{Zhang}, Y.~{Ueng}, C.~{Studer}, and A.~{Burg}, ``Artificial intelligence
  for {5G} and beyond {5G}: Implementations, algorithms, and optimizations,''
  \emph{IEEE Journal on Emerging and Selected Topics in Circuits and Systems},
  vol.~10, no.~2, pp. 149--163, June 2020.

\bibitem{Keating_Overview_2019}
R.~{Keating}, M.~{Säily}, J.~{Hulkkonen}, and J.~{Karjalainen}, ``Overview of
  positioning in {5G} new radio,'' in \emph{16th International Symposium on
  Wireless Communication Systems (ISWCS)}, Aug. 2019, pp. 320--324.

\bibitem{Chapre_2014}
Y.~{Chapre}, A.~{Ignjatovic}, A.~{Seneviratne}, and S.~{Jha}, ``{CSI-MIMO}:
  Indoor {Wi-Fi} fingerprinting system,'' in \emph{39th Annual IEEE Conference
  on Local Computer Networks}, 2014, pp. 202--209.

\bibitem{Sun_TVT_2018}
X.~{Sun}, X.~{Gao}, G.~Y. {Li}, and W.~{Han}, ``Single-site localization based
  on a new type of fingerprint for massive {MIMO}-{OFDM} systems,'' \emph{IEEE
  Transactions on Vehicular Technology}, vol.~67, no.~7, pp. 6134--6145, July
  2018.

\bibitem{Meng_LANMAN_2020}
J.~{Meng}, A.~{Sharma}, T.~X. {Tran}, B.~{Balasubramanian}, G.~{Jung},
  M.~{Hiltunen}, and Y.~{Charlie Hu}, ``A study of network-side {5G} user
  localization using angle-based fingerprints,'' in \emph{IEEE International
  Symposium on Local and Metropolitan Area Networks}, July 2020, pp. 1--6.

\bibitem{zhu05survey}
X.~Zhu, ``Semi-supervised learning literature survey,'' Computer Sciences,
  University of Wisconsin-Madison, Tech. Rep. 1530, 2005.

\bibitem{Pan_TPAMI_2012}
J.~J. {Pan}, S.~J. {Pan}, J.~{Yin}, L.~M. {Ni}, and Q.~{Yang}, ``Tracking
  mobile users in wireless networks via semi-supervised colocalization,''
  \emph{IEEE Transactions on Pattern Analysis and Machine Intelligence},
  vol.~34, no.~3, pp. 587--600, March 2012.

\bibitem{Pulkkinen_2011}
T.~{Pulkkinen}, T.~{Roos}, and P.~{Myllym{\"a}ki}, ``Semi-supervised learning
  for {WLAN} positioning,'' in \emph{Artificial Neural Networks and Machine
  Learning}.\hskip 1em plus 0.5em minus 0.4em\relax Springer Berlin Heidelberg,
  2011, pp. 355--362.

\bibitem{Bast_VTC2020}
S.~D. {Bast}, A.~P. {Guevara}, and S.~{Pollin}, ``{CSI}-based positioning in
  massive {MIMO} systems using convolutional neural networks,'' in \emph{IEEE
  91st Vehicular Technology Conference}, May 2020, pp. 1--5.

\bibitem{WSA_mMIMO_data_2020}
M.~{Gauger}, M.~{Arnold}, and S.~{ten Brink}, ``Massive {MIMO} channel
  measurements and achievable rates in a residential area,'' in \emph{24th
  International ITG Workshop on Smart Antennas}, Feb. 2020, pp. 1--6.

\bibitem{CC_2017}
C.~{Studer}, S.~{Medjkouh}, E.~{Gonultaş}, T.~{Goldstein}, and O.~{Tirkkonen},
  ``Channel charting: Locating users within the radio environment using channel
  state information,'' \emph{IEEE Access}, vol.~6, pp. 47\,682--47\,698, 2018.

\bibitem{MPCC_Deng_2018}
J.~{Deng}, S.~{Medjkouh}, N.~{Malm}, O.~{Tirkkonen}, and C.~{Studer},
  ``Multipoint channel charting for wireless networks,'' in \emph{52nd Asilomar
  Conference on Signals, Systems, and Computers}, Oct. 2018, pp. 286--290.

\bibitem{Studer_SPAWC2019}
P.~{Huang}, O.~{Castañeda}, E.~{Gönültaş}, S.~{Medjkouh}, O.~{Tirkkonen},
  T.~{Goldstein}, and C.~{Studer}, ``Improving channel charting with
  representation-constrained autoencoders,'' in \emph{IEEE 20th International
  Workshop on Signal Processing Advances in Wireless Communications (SPAWC)},
  July 2019, pp. 1--5.

\bibitem{Studer_Allerton2019}
E.~{Lei}, O.~{Castañeda}, O.~{Tirkkonen}, T.~{Goldstein}, and C.~{Studer},
  ``Siamese neural networks for wireless positioning and channel charting,'' in
  \emph{57th Annual Allerton Conference on Communication, Control, and
  Computing (Allerton)}, Sep. 2019, pp. 200--207.

\bibitem{SSMPCC2021}
J.~Deng, O.~Tirkkonen, J.~Zhang, X.~Jiao, and C.~Studer, ``Network-side
  localization via semi-supervised multi-point channel charting,'' in
  \emph{2021 International Wireless Communications and Mobile Computing
  ({IWCMC})}, 2021, pp. 1654--1660.

\bibitem{tSNE}
L.~V.~D. Maaten and G.~Hinton, ``Visualizing data using {t-SNE},''
  \emph{Journal of machine learning research}, vol.~9, pp. 2579--2605, Nov.
  2008.

\bibitem{KNN_ICC2020}
A.~Sobehy, r.~Renault, and P.~Mühlethaler, ``{CSI-MIMO}: K-nearest neighbor
  applied to indoor localization,'' in \emph{IEEE International Conference on
  Communications (ICC)}, 2020, pp. 1--6.

\bibitem{Rashdan_2020}
W.~Y. Al-Rashdan and A.~Tahat, ``A comparative performance evaluation of
  machine learning algorithms for fingerprinting based localization in
  {DM-MIMO} wireless systems relying on big data techniques,'' \emph{IEEE
  Access}, vol.~8, pp. 109\,522--109\,534, 2020.

\end{thebibliography}

\end{document}